\documentclass[9pt, twocolumn]{article}

\usepackage{graphics}
\usepackage{graphicx}
\usepackage[font=footnotesize]{caption}
\usepackage{subcaption}
\usepackage{hyperref}
\usepackage{geometry}
\usepackage{titlesec}
\titleformat{\section}
  {\normalfont\large\bfseries}
  {\thesection}{1em}{}
\titleformat{\subsection}
  {\normalfont\bfseries}
  {\thesubsection}{1em}{}

\usepackage{cite}

\title{\LARGE \bf
Reinforcement Learning Control of a Forestry Crane Manipulator
}

\author{Jennifer Andersson$^{1}$, Kenneth Bodin$^{2}$, Daniel Lindmark$^{2}$, Martin Servin$^{1,2}$ and Erik Wallin$^{1}$
\renewcommand\footnotemark{}

\thanks{$^{1}$Department of Physics, Ume\aa\ University, \{jennifer.andersson, martin.servin, erik.wallin\}@umu.se}%
\thanks{$^{2}$Algoryx Simulation AB, \{kenneth.bodin, daniel.lindmark\}@algoryx.se}
\thanks{This work has in part been supported by Mistra Digital Forest (Grant DIA 2017/14
\#6). Extractor AB has kindly provided 3D models for the Xt28 forwarder.}
}%

\begin{document}

\maketitle
\thispagestyle{empty}
\pagestyle{empty}

\begin{abstract}

Forestry machines are heavy vehicles performing complex manipulation tasks in unstructured production forest environments. 
Together with the complex dynamics of the on-board hydraulically actuated cranes, the rough forest terrains have posed a particular challenge in forestry automation. 
In this study, the feasibility of applying reinforcement learning control to forestry crane manipulators is investigated in a simulated environment.
Our results show that it is possible to learn successful actuator-space control policies for energy efficient log grasping by invoking a simple curriculum in a deep reinforcement learning setup. Given the pose of the selected logs, our best control policy reaches a grasping success rate of 97\%. 
Including an energy-optimization goal in the reward function, the energy consumption is significantly reduced compared to control policies learned without incentive for energy optimization, while the increase in cycle time is marginal.
The energy-optimization effects can be observed in the overall smoother motion and acceleration profiles during crane manipulation. 

\end{abstract}

\section{INTRODUCTION}

Recent advances in machine learning (ML) in general, and reinforcement learning (RL) in particular, have inspired progress in the development of intelligent systems in the context of robotic manipulation \cite{ sunderhauf2018limits, ibarz2021}. 
These tasks often require multiple-skill acquisition in high-dimensional and dynamically complex settings. 
RL has the advantage of not requiring large static datasets and direct supervision, while enabling end-to-end learning through environmental exploration and experience. 
Successful applications of RL to robotic manipulation problems include robotic grasping \cite{Kalashnikov2018QTOptSD}. 
Though many challenges remain, 
synergies between RL and robotics have the potential to accelerate autonomous systems development in the automotive and robotics industry. 
So far, most RL research in this area has been limited to learning individual skills or performing generic manipulation tasks. 
However, the avenue for leveraging this technology is much broader. 
The current work aims to extend the application area of RL control to heavy equipment in unstructured environments, using deep RL techniques to fully automate the log-grasping motion of a forestry crane manipulator. 

\begin{figure}[t]
      \centering      
      \includegraphics[trim=0mm 0mm 0mm 30mm, clip, width=0.47\textwidth]{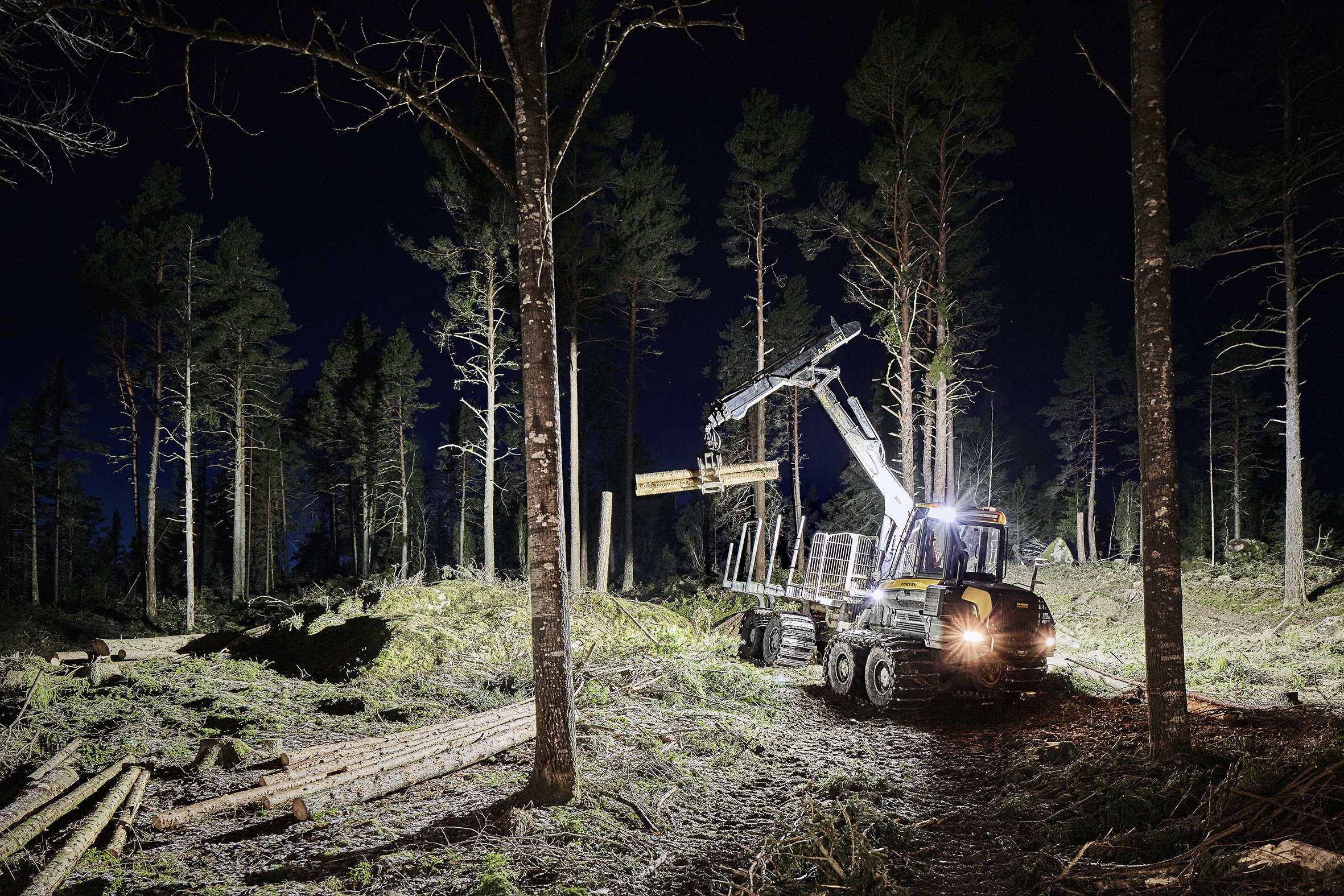}
      \caption{A forwarder grasping and loading logs in a forest using a crane manipulator. Image courtesy of Holmen.}
      \label{forwarder}
      \end{figure}

The forest environment is notably unstructured and dynamically complex, see Fig.~\ref{forwarder} for an example.
This has contributed to the comparatively slow automation progress in the forest industry over the past decades, but advances in machine learning has rekindled ambitions for end-to-end automation to play a significant role in the future of forestry. 
Forestry machines permeate the entire logging process, from the felling of trees to the cutting, sorting and transportation of timber out of the forest harvesting site. 
The key equipment carried by these machines is hydraulic, kinematically redundant manipulators used for monotonous manipulation tasks. 
The system is underactuated, which increases control complexity compared to most small-scale manipulators. 
Despite widespread automation in industry today, forestry machines have remained primarily manually operated. 
For a human operator, manual control of a forestry crane manipulator can be a both mentally and physically exhausting task, requiring counterintuitive coordination of several actuators for many hours straight and exposing the operator to harmful whole-body vibrations \cite{lahera2019}. 
This can have severe long-term health implications 
and the need for increased automation in the forestry context is therefore substantial.

Semi-automation of forestry machines has been successfully explored before, increasing productivity and reducing the immediate workload on the operator \cite{hansson2010, manner2017}. 
In contrast to methods previously adopted in forestry automation research, the RL framework has shown potential for intelligent and independent end-to-end learning of complex tasks in simulated environments, completely removing the need for analytic low-level controllers. 
If these techniques can be used to achieve full automation of essential parts of the boom cycle in simulation, this may serve as a stepping stone towards full automation of physical forestry machines, or at the very least accelerate semi-automation efforts. 

To the best of our knowledge, RL control of forestry crane manipulators is a topic previously not touched upon in machine learning or robotics research. 
Our work constitutes an initial attempt at investigating the feasibility of adopting this approach to learn successful actuator-space control policies for single-log grasping using a 6 DoF, kinematically redundant forestry crane manipulator. 
Given the Cartesian position and orientation of the selected log, the learned control policies map task-space goals directly to actuator-space commands. 
Training and testing is conducted in a simulated environment, and curriculum learning is used to deal with the longstanding challenge of sparse environmental feedback characteristic of robotic grasping problems. This approach removes the need for conventional trajectory planning and extensive reward shaping. 
Secondly, we investigate the policy response following inclusion of an energy-optimization goal in the reward function. 
In addition to effects on performance, we analyse differences in crane behavior and acceleration profiles compared to unoptimized policies. 
Finally, to investigate simulation to reality transferability, we analyse the sensitivity and robustness of successful control policies exposed to environmental disturbances and uncertainties in the observation and parameter space, respectively. 

\section{BACKGROUND}

\subsection{Robotic Control in Forestry}

Most forestry operations involve maneuvering heavy vehicles over rough terrain, and manipulating the unstructured environment with the end-effector of a hydraulically actuated crane, as elaborated on in \cite{westerberg2014}.
In this study, we focus on forwarding, but the automation challenges are similar for harvesting and thinning.  
Forwarding is the operation of loading and transporting logs from a felling site to a nearby forest road, where the logs are unloaded for further transportation by road vehicles. 
The responsibility of the forwarder operator includes both low-level control in terms of vehicle maneuver and crane manipulation, and high-level navigation, planning, task coordination and execution optimization.
The loading cycle consists of extracting and slewing the crane, guiding an open grapple to a selected log from above. 
The log is gripped and loaded by closing the grapple, slewing back and retracting the crane to bring the log to the forwarder's load bunk, while avoiding collisions of any kind.
This requires object detection and pose estimation, as well as strategic selection of logs and grasping configuration. In addition, operation of the crane demands motion planning and control answering to the physical limitations of the crane depending on the vehicle position and inclination. 
A typical crane reaches up to 10 metres, has four links and hydraulic actuators delivering a lifting torque of 100 kNm. 
The grapple and rotator add two degrees of freedom to actuate closing and opening as well as the axial rotation for aligning the grapple with a log or the load bunk.

The crane dynamics is prone to oscillations. This complicates the grasping process, and causes excessive wear and discomfort to the operator.
Operators undergo extensive training, learning to operate the forwarder as time- and energy efficiently as possible.
Still, more than 80\% of the operator's active time is devoted to controlling the crane \cite{dvorak2008}.
Thus, crane manipulation is a natural starting point for research motivated by forestry automation.

It is found that, on average, machine operators are capable of using only 20\% of the maximum velocity of the crane during operation \cite{lahera2019}. 
According to Morales et al. \cite{Morales2015}, time-efficiency could increase by at least three-fold if done by an autonomous control system. 
Crane-tip Cartesian control and semi-autonomous functions improve the performance of inexperienced operators and reduce the workload on experienced operators \cite{manner2017,hansson2010,westerberg2014}. 
However, previous research on motion planning and crane control \cite{OrtizMorales2014} has disregarded grapple control and the task of log grasping.
Moreover, there are only a few publications that deal with machine vision for forestry robotics, e.g. detection and pose estimation of logs \cite{park2011}
or tree stems \cite{nevalainen2020}.
This is recognized as a difficult problem in the forestry environment, which is characterized by high variability, object occlusion, and presence of moisture and particles. 
On the other hand, production forests are monitored using aerial lidar mapping with steadily increasing resolution \cite{white2021}
and the GPS position of felled logs are registered with a precision of a few meters \cite{talbot2018}. Combining these data with on-board sensors, it is conceivable that logs can be located with a resolution comparable to the grapple.

\subsection{RL Preliminaries}

An RL problem can be formalized as a \emph{Markov decision process}; a mathematical framework for sequential decision making in stochastic state-transition systems. 
At a given discrete time step $t$, the system is in state $s_t \in \mathcal{S}$ and the agent makes an observation $o_t \in \mathcal{O}$ of the environment. 
Performing an action $a_t \in \mathcal{A}$ according to the policy distribution $\pi(a|s)$, the agent receives an immediate scalar reward $r_t(s_t, a_t)$ according to the specified reward function $\mathcal{R}(s,a)$. 
The goal of RL algorithms is to find the optimal policy $\pi^*(a|s)$ such that the agent takes the optimal action at any given state in order to maximize the expected return.

Here, the deep RL approach involves parameterizing the policy $\pi$ as a neural network $\pi_\theta$ with parameters $\theta \in \Theta$.
The resulting policy approximator outputs a vector of actuator-space motor control signals at each time step. 

\subsection{Curriculum Learning in RL}

The concept of transfer learning (TL) \cite{zhu2020} has shown potential to solve high sample complexity issues symptomatic of many RL applications. 
In TL, the agent learns to master a simpler source task and uses the acquired knowledge to bias learning on the original target task. 
Ideally, the TL approach augments and speeds up learning on the target task, and is especially beneficial in context where learning is inherently slow, often occurring in settings exhibiting sparse rewards and feedback delay. 
In the context of log grasping, the adverse effect of sparse rewards on the learning process is imminent.
Such reward signals only require a definition of success in relation to the goal, allowing the agent to find the optimal solution given the environmental constraints.
The drawback is that a significant part of training is devoted to endless exploration with rare to no feedback from which the agent can learn.
This motivates us to deploy a curriculum learning-based \cite{narvekar2020} solution, in which a sequence of source tasks is carefully combined to form a curriculum of lessons, easing learning of the target task by repeated use of TL.    
\section{SYSTEM OVERVIEW}

\subsection{Delimitations}

The grasping subtask of the forwarder operator is very complex in the real forest environment.
To this end, our initial work is limited to RL control of the single-log grasping motion of a forestry crane manipulator mounted on a static machine fixed on a horizontal surface.
The task of returning logs to the load bunk is omitted in the present study.
Moreover, the existence of an external perception system is assumed, and the agent observes the Cartesian position and orientation of the selected log directly.

\subsection{Virtual Forwarder} 

\begin{figure}[!t]
      \centering
      \includegraphics[width=0.9\columnwidth]{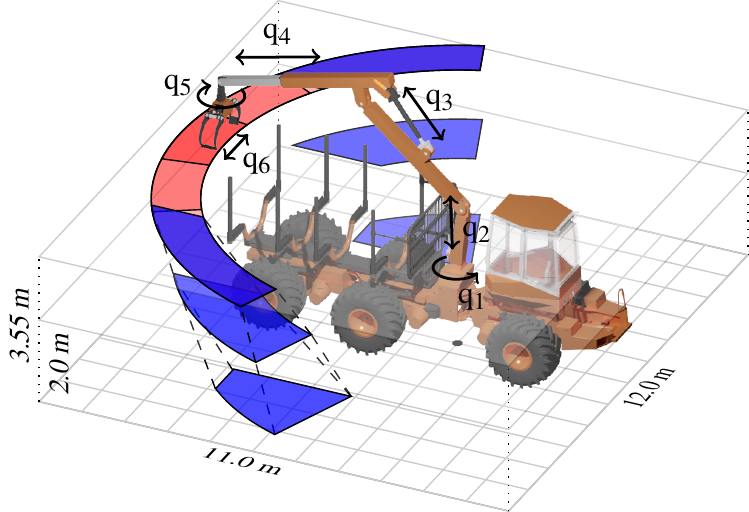}
      \caption{The forwarder and the selected log-position areas.
      At initial lessons, the log is placed close to the initial position of the grapple (red). 
      As the agent progresses through the curriculum, the log-position area changes (blue), incrementally increasing the difficulty of the task as the artificial plane approaches the ground plane.}
      \label{curriculum}
\end{figure}

Our agent controls a 3D simulation model of a six-wheel forwarder equipped with a kinematically redundant hydraulic manipulator controlled by 6 actuated DoF. The model, depicted in Fig.~\ref{curriculum}, is a slightly modified version of the Xt28 concept forwarder \cite{gelin2020}, consisting of 52 rigid bodies and 60 constraints. It has a total mass of 16.800 kg, whereof the crane stands for 16 bodies, 20 constraints, and 2.000 kg.
The smallest component weighs 6 kg, yielding a mass ratio above 1000 to the vehicle.
The crane reach is 7 m.

The position of the boom tip at the end of the telescope, where a rotator grapple is attached, depends on the rotational angles of three revolute joints and one linear displacement position. 
A revolute joint, $q_1$, controls the slewing motion of the crane pillar. 
The inner and outer booms are controlled by two actuated prismatic joints, $q_2$ and $q_3$, representing one hydraulic cylinder each. 
A motorized prismatic joint, $q_4$, controls translation of the telescope along its axis. 
The orientation of the grapple along the rotator axis is controlled by the actuated revolute joint $q_5$, and a final prismatic joint, $q_6$, controls the opening-closing motion of the grapple. 
This leaves the rotator grapple with two degrees of freedom that are not actuated. This is important for focusing the stress in the direction of lifting, consequently allowing the grapple to swing freely and denying direct control of the grapple claws.

On conventionally operated forestry cranes, joystick signals control the hydraulic flow rates of the cylinders and rotator. A diesel engine powers a hydraulic pump applying fluid pressure within the system. In our case, the diesel engine is not modelled, and each actuated joint is equipped with a linear or rotational motor internally controlled to reach and hold a given target velocity, while obeying current dynamics and constraints, such as maximum motor force, torque or speed. 
Each joint may also have secondary constraints in terms of range limits or a lock, enforcing a fixed joint position. 
Physical parameters of the virtual forwarder are estimated to emulate the force and speed limitations of a real crane, but have not been validated by domain experts. 
This does not reduce the generality of the problem.
The initial crane configuration centres the grapple above the load bunk and the initial angle of each actuator are slightly perturbed to prevent overfitting. Logs are modelled as uniform cylinders of length 3 m (1.5 m during training, to encourage grasping near the log C.o.M), radius 0.08 m and mass 50 kg. 
The initial C.o.M position is drawn from a uniform probability distribution within the selected area depicted on the nearside of the vehicle in Fig. \ref{curriculum}. The orientation in the horizontal plane is drawn from a similar distribution.

\subsection{Learning Environment}
The Unity 3D simulation platform is used as a training environment, interfaced with the ML-Agents Toolkit \cite{unity2020} and the physics engine AGX Dynamics \cite{agx}. 
Owing to its block-sparse direct-iterative solver and symmetry-preserving variational stepper, AGX Dynamics supports nonsmooth multibody dynamics with frictional contacts and large mass-ratio mechanisms with high numerical precision and speed. This is demanded by the current application. 
We use a fixed simulation time step of 20 ms and each training session is run for at least 35 million time steps ($\sim 200$k episodes) using eight parallel environments. 
The observation space consists of the Cartesian log position and orientation, as well as the current state of each actuator in terms of angle, speed and applied motor torque. Each observation is normalized based on the running mean and standard deviation of previous observations, and eight stacked observations are processed at each time step. 
The agent is controlled by action signals corresponding to the target speed of each motor, respecting its specific velocity range.
The instantaneous change in target speed per time step is limited to $1/30$ of the maximum speed of the motor.

Our policies are parametrized by a feedforward neural network and optimized using the state-of-the-art on-policy algorithm \emph{Proximal Policy Optimization} (PPO) \cite{schulman2017}. We use a network with three fully-connected hidden layers each comprised of 256 neurons, and a linearly decaying learning rate of $\alpha=0.001$. 
The maximum episode length is 2000 time steps, and the agent makes a decision every other time step.  
For PPO, we use the clipping parameter $\epsilon=0.3$, the entropy regularization coefficient $\beta = 0.01$, the GAE parameter $\lambda=0.95$ and the discount factor $\gamma=0.995$. 

\subsection{Reward Structure \& Curriculum} 
Succeeding to secure the selected log in its grapple, the agent receives a high sparse reward and the episode is terminated. 
For our purposes, a successful grasp occurs when the log is lifted from the ground, enclosed by both grapple claws in a closed position. 
Incentive for energy optimization is included by scaling the reward inversely with the total energy consumed by the actuators $q_1$ to $q_4$, excluding the energy consumed by the grapple actuators.
The energy consumption is defined as the work exerted by the actuators up until the point of grasping initiation, assuming no energy recuperation. 
This reward structure enables optimal policy search, but does nothing to prohibit unwanted behavior allowed for by the model.
Thus, episodes terminate with a zero-return if maximum motor torque is applied at the range limit of $q_1, q_2$ or $q_3$, or if any part of the crane collides with the load bunk.
This increases learning speed and avoids contact-heavy computations.

An additional reward signal, increasing exponentially with decreasing distance between the boom tip and the log, is provided. This guides the agent at the start of the learning process, but is negligible once any grasping behavior has been learned. This reward signal increases with decreasing $q_4$ speed in close proximity to the log, and vanishes if $q_4$ hits its range limits or if there is significant deviation between the orientation of the grapple and the orientation of the log. This is effective when the goal is to navigate the boom tip to, and remain at, a predefined grasping position. Extending the goal to include grasping, however, the exploration space becomes too large for this reward signal to suffice. When the agent finally learns to navigate the grapple to the log, it has learnt to discard the opening-closing motion of the grapple. Alleviating this challenge, we deploy a straightforward curriculum in which the distance between the log and the grapple increases incrementally. 
This is accomplished by adjusting the height of an artificial ground plane carrying the selected log. This avoids the introduction of bias induced from guiding the crane using analytic motion control.  
Throughout the first four lessons, the artificial plane is placed directly beneath the initial grapple position, with the selected log-position area expanding according to Fig. \ref{curriculum}. 
Between succeeding lessons, the artificial plane is lowered in intervals of 0.1 m. The grapple reach varies with the target height, and the log-position area is adjusted accordingly throughout the curriculum. 
At the final lesson, the artificial plane merges with the true ground plane and the agent continues training on the target task.

A grasping success rate of 30\% over the preceding 20 episodes is required to proceed to the next lesson, preventing overfitting early in the curriculum. 
Our goal is to quickly reach ground level to train on the target task, motivating us to decrease the problem complexity of early lessons by disabling collisions between the artificial plane and the grapple claws. 

\section{RESULTS \& ANALYSIS}

\subsection{Performance Evaluation}

\setlength\tabcolsep{3pt}
\begin{table}[t]
\captionsetup{font=scriptsize}
\caption{Comparison of performance between Policy A-D in terms of success rate, average cycle time, relative energy consumption and training time. Policy B, C and D are energy optimized.}
\label{table_performance}
\begin{center}
\begin{footnotesize}
\begin{tabular}{|c||c||c||c||c|}
\hline
Policy & Success Rate & Time (s) & Energy & Training Steps\\
\hline
A & 0.97 & 3.6 & 1 & 35e6\\
\hline
B & 0.81 & 4.6 & 0.39 & 80e6\\
\hline
C & 0.84 & 4.9 & 0.22 & 80e6\\
\hline
D & 0.93 & 4.0 & 0.32 & 80e6\\
\hline
\end{tabular}
\end{footnotesize}
\end{center}
\end{table}

\begin{figure}[t]
      \centering
      \includegraphics[width=1.6in]{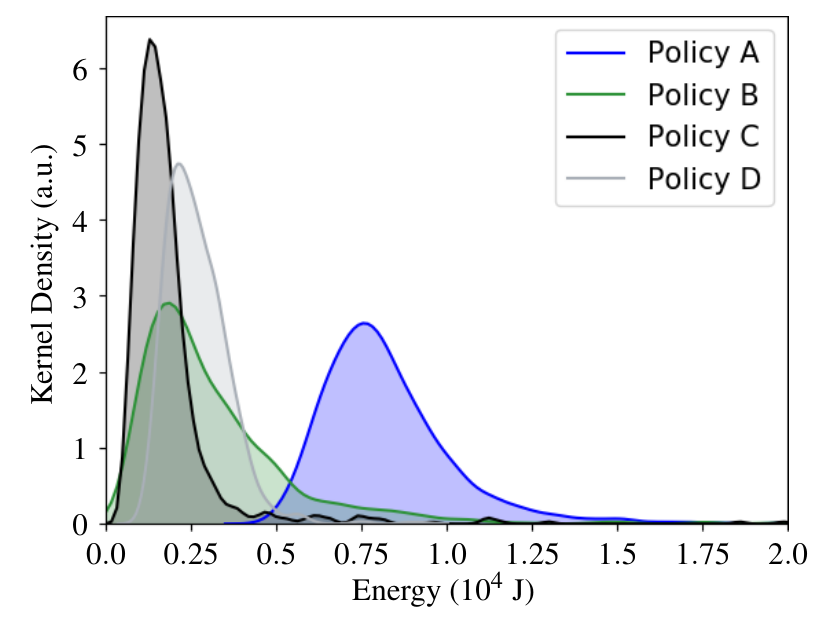}
      \caption{Total energy consumed by $q_1$ to $q_4$ up until the point of grasping. Energy distributions over 1000 episodes are shown for Policy A-D.}
      \label{energy}
   \end{figure}

\begin{figure}[!t]
  \centering
  \includegraphics[width=0.49\columnwidth]{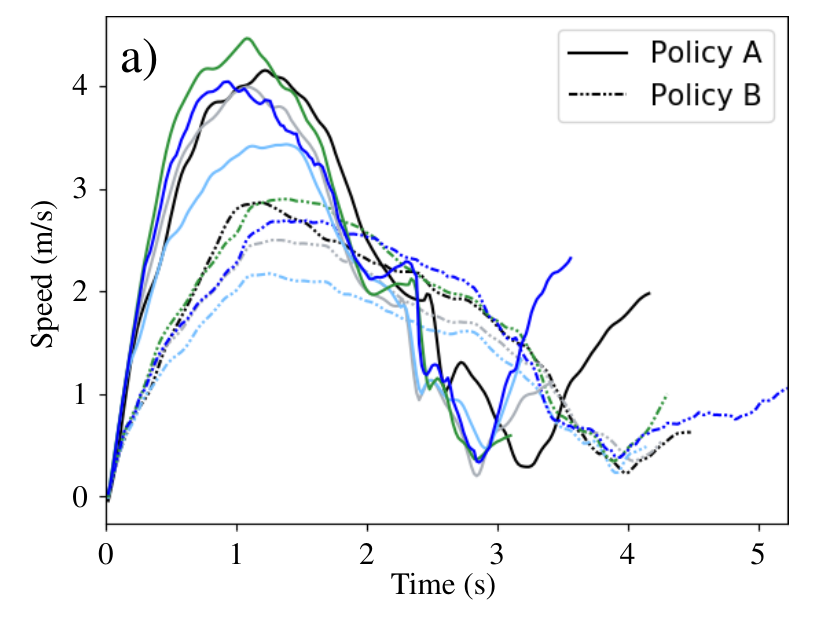}
  \includegraphics[width=0.49\columnwidth]{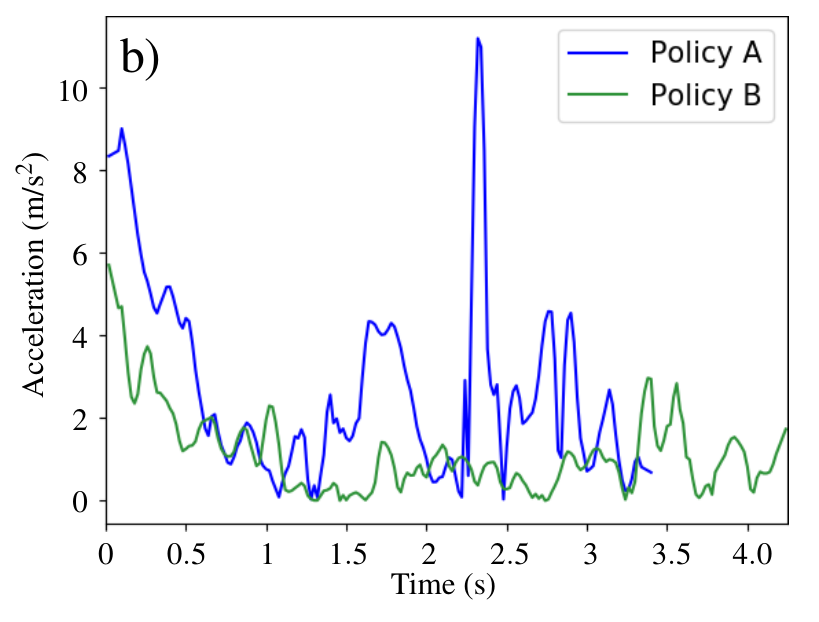}
  \caption{Boom tip trajectory profiles produced by Policy A and B over a) five boom cycles for logs selected in different parts of the grasping area and b) a randomly selected boom cycle.}
  \label{acceleration_speed}
\end{figure}

\begin{figure*}
   \includegraphics[trim={20mm 0 0mm 0},clip,width=0.14\textwidth]{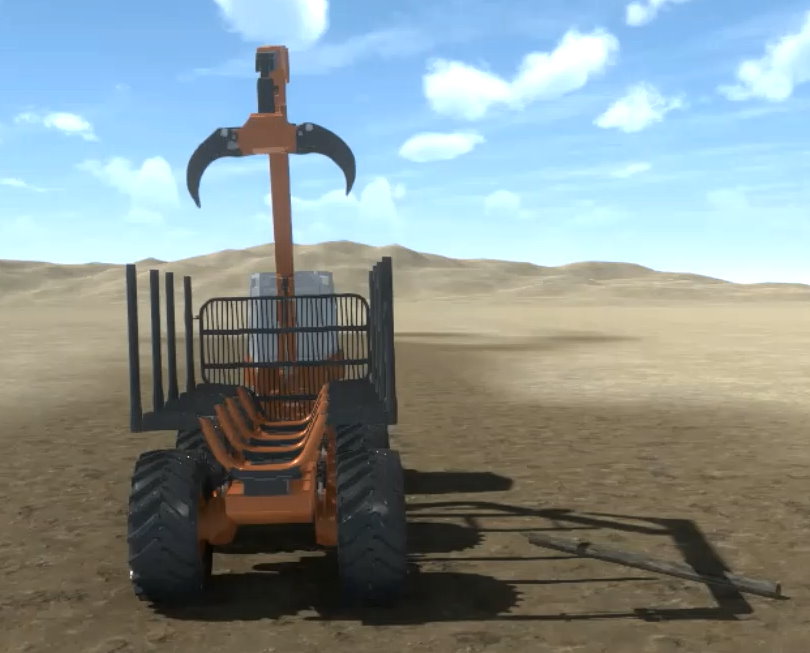}\hspace{-1mm}
   \includegraphics[trim={20mm 0 0mm 0},clip,width=0.14\textwidth]{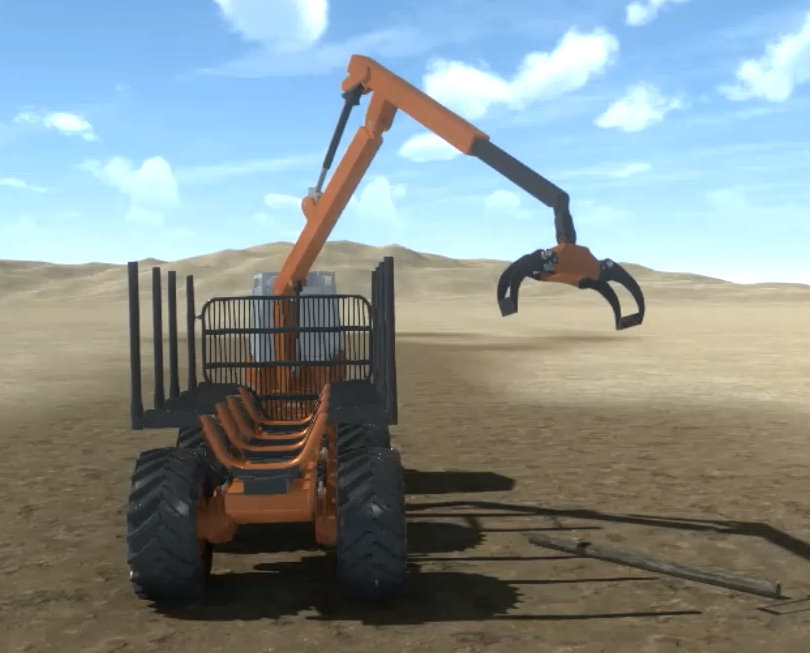}\hspace{-1mm}
   \includegraphics[trim={20mm 0 0mm 0},clip,width=0.14\textwidth]{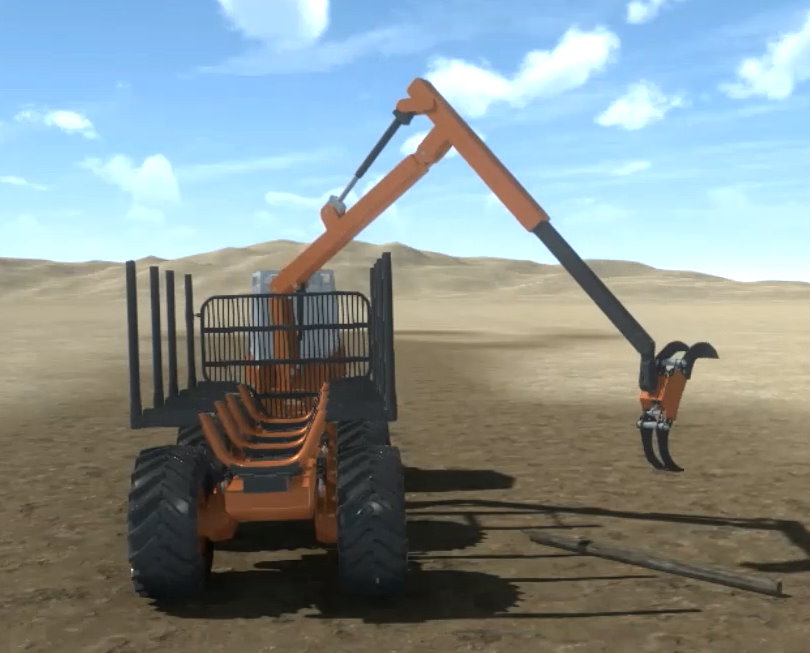}\hspace{-1mm}
   \includegraphics[trim={20mm 0 0mm 0},clip,width=0.14\textwidth]{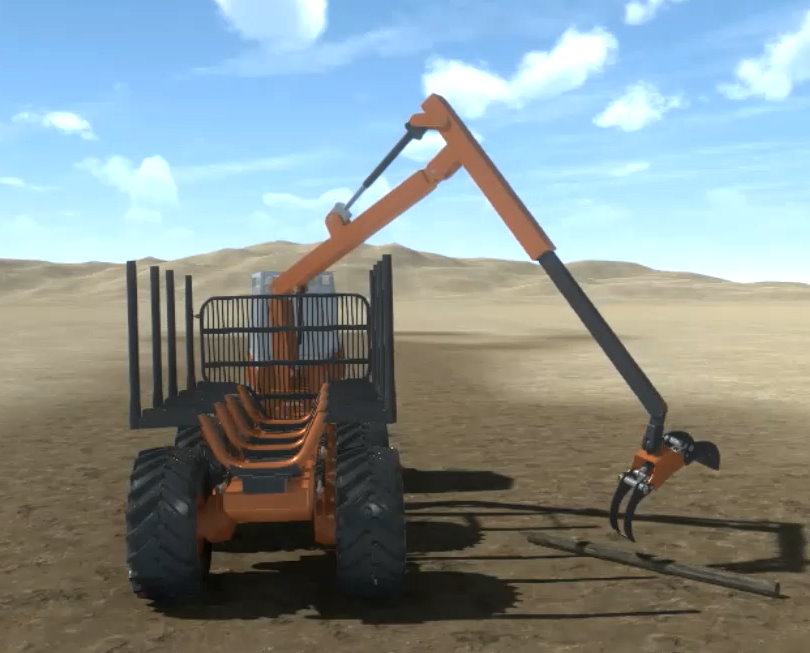}\hspace{-1mm}
   \includegraphics[trim={20mm 0 0mm 0},clip,width=0.14\textwidth]{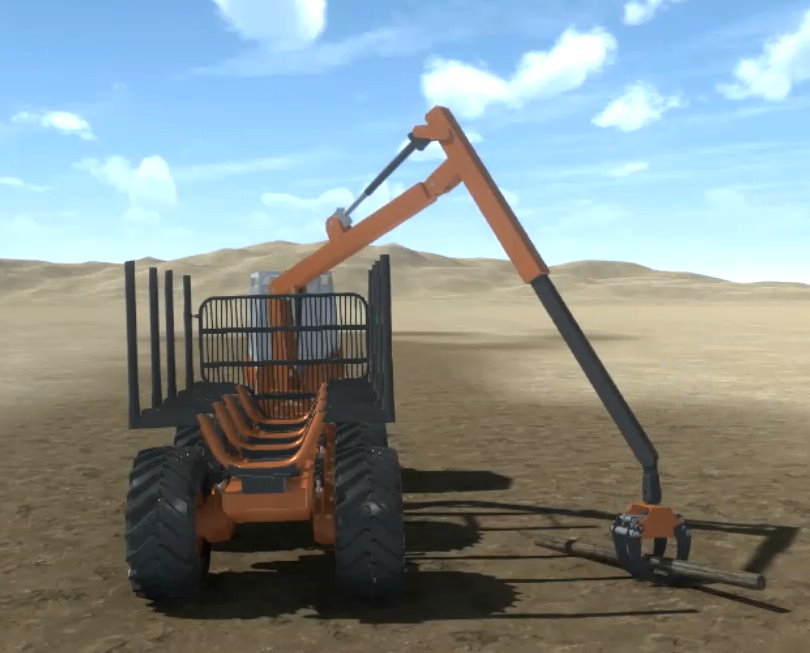}\hspace{-1mm}
   \includegraphics[trim={20mm 0 0mm 0},clip,width=0.14\textwidth]{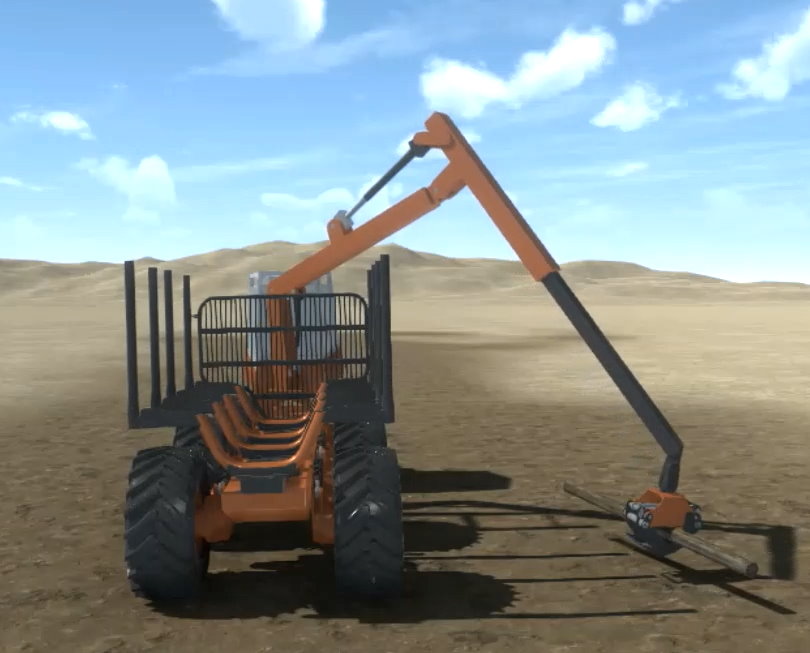}\hspace{-1mm}
   \includegraphics[trim={20mm 0 0mm 0},clip,width=0.14\textwidth]{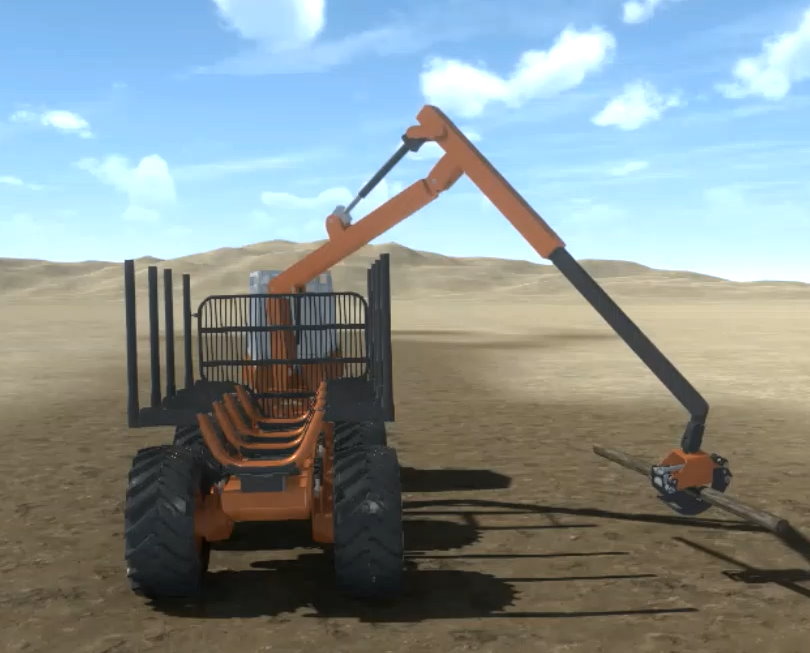}
   \caption{Grasping sequence using Policy A showing snapshots taken 0, 1, 1.5, 2, 2.5, 3 and 4 s into the boom cycle.}
   \label{sequence}
 \end{figure*}

For each control policy, the grasping success rate over 1000 consecutive episodes is recorded. The best policy optimized without incentive for energy optimization reaches a near perfect evaluation success rate of 97\%. This policy is referred to as Policy A. Fig. \ref{sequence} shows a grasping sequence in the simulation environment using this policy. Three independent policies (B, C and D) are learned using the reward function including an energy-optimization goal. Table \ref{table_performance} compares the performance of these policies, and the generated grasping behaviors are demonstrated in the supplementary videos. In the given training time, our best energy-optimized policy reaches a success rate of 93\%. The energy reduction including incentive for energy optimization is substantial compared to Policy A, with the total energy consumption on average reducing by 61\% (Policy B), 78\% (Policy C) and 68\% (Policy D). Fig. \ref{energy} compares distributions of the total energy consumed during successful boom cycles using each policy.
The most prominent energy reduction effect can be observed in the smoother trajectory profiles produced by the energy-optimized policies. This is illustrated in Fig. \ref{acceleration_speed}a, showing the boom tip speed profiles over five boom cycles for Policy A and B. 
The boom tip acceleration profile during one of these cycles is illustrated in Fig. \ref{acceleration_speed}b. 
We observe a generally lower boom tip speed and a significant jerk reduction using the energy-optimized policies. 
As a result of these effects, the grapple oscillations are significantly reduced and the average cycle time increases by 11\%-36\% compared to Policy A, as seen in Table \ref{table_performance}.   
These cycle times are comparable to those obtained by manual operators.

\begin{figure}[!t]
  \centering
  \includegraphics[width=0.49\columnwidth]{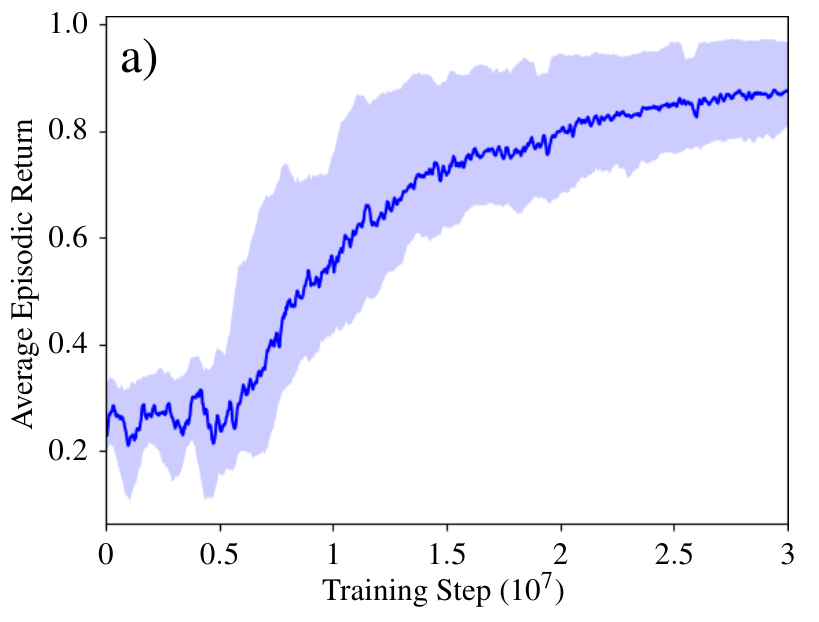}
  \includegraphics[width=0.49\columnwidth]{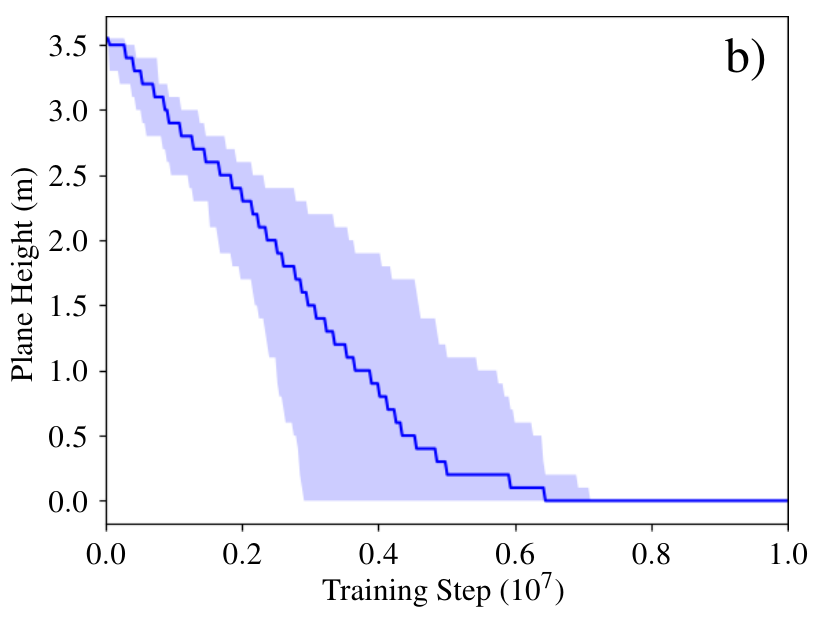}
  \caption{Learning curves averaged over five random seeds in terms of a) mean episodic return corresponding to the grasping success rates during training and b) evolution through the curriculum during training as a function of artificial plane height. Training is initialized from a policy trained on the first four lessons, and training starts at the final lesson with the initial artificial ground plane height.}
  \label{learning_curve_lesson_curve}
\end{figure}

Overall, the training process is stable using our approach. 
The energy-optimized policies require longer training time to reach comparable success rates, and may converge to higher success rates if allowed more time to train. 
Fig. \ref{learning_curve_lesson_curve}a shows learning curves for policies optimized without incentive for energy optimization. The mean success rate of these models is 90\%. 
Fig. \ref{learning_curve_lesson_curve}b shows the training process in terms of evolution through the curriculum. 
The deviation in training time before the initiation of learning on the target task can amount to several million training steps. 
This may or may not impact the success rate to which the policy converges in a given number of training steps. 
In our case, the superior policy in terms of success rate (Policy A) reaches the final lesson significantly faster than all other policies.

Due to the redundant crane kinematics, successful grasping poses can be reached from an infinite number of crane configurations. 
Analysing the end-position of all actuators upon grasping, two primary solutions can be observed among our models. These can be distinguished by the respective use of $q_4$, where one primary solution involves using the crane telescope to maximum or close to maximum capacity, practically removing one of the redundant degrees of freedom and adjusting $q_2$ and $q_3$ to reach the grasping pose.
The energy-optimized policies exhibit similar primary solutions in terms of crane behavior, and
displays a distinct second solution positioning $q_2$ with low variance across different boom cycles, adjusting $q_3$ and $q_4$ to reach the grasping pose. The latter is true for Policy B and C, whereas Policy D uses $q_4$ heavily.
In this context, using $q_2$ can allow most of the crane mass to fall with gravity to reduce energy consumption. 
Operators are instructed to use the maximum capacity of $q_4$, similar to one of the approaches favored by our policies.  
This is considered optimal maneuver over the course of an entire loading cycle, but is not necessarily the optimal behavior under our delimitation. 
A particularly interesting indication is that our policies are able to take advantage of grapple oscillations to perform grasping, and use the grapple claws to nudge the log to align better with the grapple orientation.
This is typical of experienced operators.  

No strong correlation between failed grasping attempts and the log position can be observed. Instead, failed attempts occur due to occasional miscoordination between actuators, and depends on the link activation profile favored by the policy. For example, for policies not grasping from above, i.e. favoring grasping under non-symmetric motion of the grapple claws relative to the log, small deviations in link activation may lead to the grapple pushing the log out of reach. 
Another example is found in Policy B, which relies on extensive use of $q_2$ and runs a greater risk of failing due to collisions between the grapple and the load bunk.     

It is worth noting that the redundant kinematics and the complexity of the problem makes it difficult to produce theoretically optimized policies, and all energy-optimized policies show significant differences in link activation. Policy B achieves significant energy reduction compared to Policy A, despite often exhibiting a slight back-and-forth motion of $q_4$ during the course of an episode. 
Policy C manages the same while generating a behavior in which the crane lowers the outer boom upon grasping initiation, effectively moving the log towards the load bunk. 
This exemplifies non-optimal behavior generated by the energy-optimized policies using the current sparse reward setup.

\subsection{Sensitivity Analysis}

Our policies are exclusively trained to grasp logs from a static vehicle. To analyse the robustness in a more realistic setting, we record the decrease in success rate performing the identical log-grasping task from a dynamic vehicle. The added flexibility, for example between the tires and terrain, induces additional crane oscillations unforeseen by the agent. 
The sensitivity is highly dependent on the specific link activation profiles. The highest success rate is recorded for Policy A, which maintains 83\% of its original success rate. 
The energy-optimized models are less robust, maintaining $33\%$ (Policy B), $53\%$ (Policy C) and $42\%$ (Policy D).   
Failed grasping attempts following Policy B is a consequence of lowering $q_2$ at an early stage of the boom cycle, which frequently results in collisions between the grapple and the load bunk when the crane is exposed to increased oscillations. 
The reduced success rates of Policy C and D are often tied to early closing of the grapple claws. 
It should be noted that this robustness can likely be increased if vehicle dynamics is included in the training process. 

Moreover, we analyse the policy robustness to inclined terrains by recording success rates with the forwarder and log placed on an uphill 17.6\% slope. Policy A is the most robust, maintaining 51\% of the original success rate. The energy-optimized models, which rely more on gravity during operation, maintain between 25\% and 45\% of the original success rates. This suggests some generalizability, but clearly demonstrate the need for training in more unstructured environments prior to deployment in real production forests.

No strong correlation between the grapple orientation and the orientation of the log can be observed in the resulting behavior of our policies. This is reasonable, as the grapple can be used to adjust the log orientation, and the grasping range of the grapple is large compared to the size of the log. High robustness to small observational uncertainties is therefore expected, and is important for prospects of future policy transfer between simulation and reality with feasible sensor accuracy. 
To investigate this, we analyse sensitivity to observational uncertainty in the Cartesian position of the log. Confining the observed position to the surface of a sphere of one log-radius around the true position, Policy A-D stays within 98\% of the original success rates. 
For policy A and B, the success rates do not decrease. 
Doubling the radius, at least 90\% of the original success rates are maintained for the energy-optimized policies, with Policy A remaining robust to 97\% of the original success rate. 
Uncertainty in log orientation yields similar robustness for Policy A, C and D, with the maximum decrease in success rate amounting to 3 percentage points (Policy D) when the observation deviates from the true log orientation by $\pm 10$ degrees in the horizontal plane. Policy B is slightly less robust, maintaining only 85\% of its original success rate. 
These results are encouraging, however, as robustness to this level of uncertainty is expected to exceed that demanded by the accuracy of available sensor technology.
Similar robustness to mass uncertainty is observed, with all policies maintaining at least $96\%$ of the original success rates under a 5\% increase in mass of each crane body. 
This shows potential for model deployment to physical machines with realistic uncertainty in crane mass estimation, inferring a possibility to transfer identical policies to multiple physical machines.

\section{CONCLUSIONS}
Using curriculum learning to solve the sparse reward log-grasping problem, we show that RL control can generate high success rates for single-log grasping using a forestry crane manipulator. 
The best control policy reaches a grasping success rate of 97\%.
Simply scaling the reward signal, it is possible to achieve significant reduction in the total energy consumed during crane manipulation, while maintaining a high success rate. 
This leads to smoother trajectory profiles compared to control policies learned without incentive for energy optimization. 
These results are important, as jerkiness poses a particular challenge to robotic control and efficient automation.
The best energy-optimized policy reaches a grasping success rate of 93\%. Our policies are largely robust to observational uncertainties, and the robustness to environmental disturbances is encouraging for future research focusing on RL control of forestry crane manipulators in more unstructured environments.

\section{SUPPLEMENTARY MATERIAL}
Supplementary material, including videos, can be found at {\small \url{http://umit.cs.umu.se/rlc_forestry_crane/}}.

\addtolength{\textheight}{-2cm}   
                                  




\end{document}